\documentclass[10pt,twocolumn,letterpaper]{article}

\usepackage{cvpr}
\usepackage{times}
\usepackage{epsfig}
\usepackage{graphicx}
\usepackage{amsmath}
\usepackage{amssymb}

\usepackage[boxed]{algorithm2e} 

\graphicspath{{./}{./figs/}}

\usepackage[pagebackref=true,breaklinks=true,letterpaper=true,colorlinks,bookmarks=false]{hyperref}

\cvprfinalcopy 

\def\cvprPaperID{1553} 

\newcommand{\hamed}[1]{{ #1}}
\newcommand{\ignore}[1]{}

\ifcvprfinal\fi
\begin{document}

\title{DeepCAMP: Deep Convolutional Action \& Attribute Mid-Level Patterns}

\author{
Ali Diba$^{1}$\thanks{A. Diba and A.M.Pazandeh contributed equally to this work},
 Ali Mohammad Pazandeh$^{2}\footnotemark[1]$,
 Hamed Pirsiavash$^{3}$,
 Luc Van Gool$^{1,4}$\\ \\
$^{1}$ESAT-PSI, KU Leuven \qquad  $^{2}$SUT \qquad $^{3}$UMBC \qquad $^{4}$CVL, ETH Zurich \\
$^{1}${\tt\small firstname.lastname@esat.kuleuven.be} \quad $^{2}${\tt\small pazandeh@ee.sharif.edu} \quad $^{3}${\tt\small hpirsiav@umbc.edu}\\
\and
}	

\maketitle
\begin{abstract}
The recognition of human actions and the determination of human attributes are two tasks that call
for fine-grained classification. Indeed, often rather small and inconspicuous objects and features 
have to be detected to tell their classes apart. In order to deal with this challenge, we propose a 
novel convolutional neural network that mines mid-level image patches that are sufficiently 
dedicated to resolve the corresponding subtleties. In particular, we train a newly designed CNN (DeepPattern)
that learns discriminative patch groups. There are two innovative aspects to this. On 
the one hand we pay attention to contextual information in an original fashion. On the other hand, 
we let an iteration of
feature learning and patch clustering purify the set of dedicated patches that we use. We validate 
our method for action classification on two challenging datasets: PASCAL VOC 2012 Action and Stanford 
40 Actions, and for attribute recognition we use the Berkeley Attributes of People dataset. Our 
discriminative mid-level mining CNN obtains state-of-the-art results on these datasets, without a 
need for annotations about parts and poses.
   
\end{abstract}

\section{Introduction}
\label{sec:intro}
\begin{figure}[t]
 \centering
\includegraphics[width=250pt]{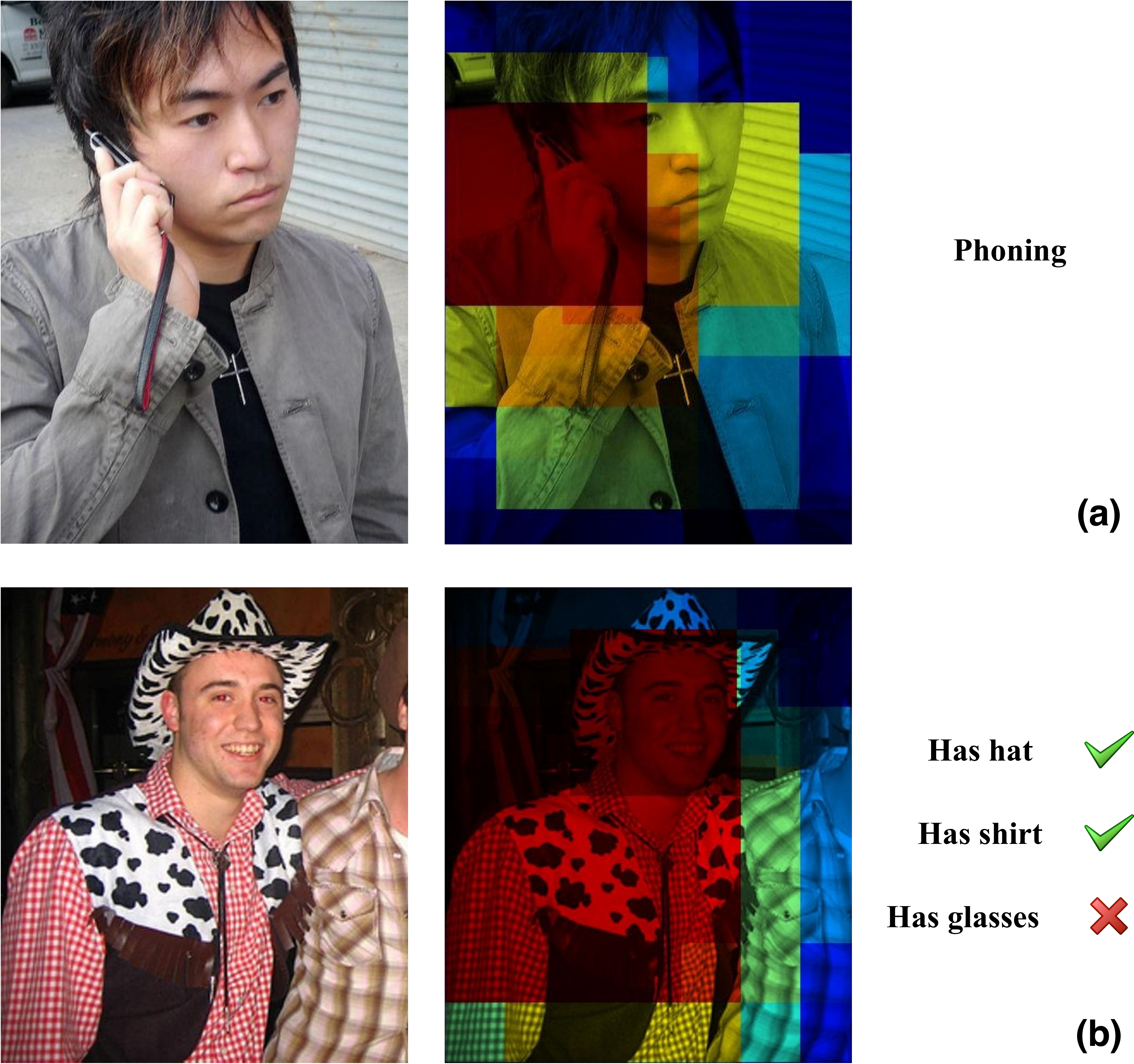}
 \caption{Mid-level visual elements: discriminative descriptors of human actions and attributes. 
          Our method discovers visual elements which make discrimination between human body parts or attributes or interacted objects
          . (a) shows the scores and classification results in the action classification task (b) shows discrimination scores of elements in the attributes classification task by color and shows final results of classification.}
 \label{fig:1}
 \end{figure}

Mimicking the human capability to understand the actions and attributes of people is very challenging. 
Lately, deep neural networks have strongly increased the capacity of computers to recognize objects,
yet the analysis of human actions and attributes is lagging behind in terms of performance. These 
are a kind of fine-grained classification problems, where on the one hand possibly small patches 
that correspond to crucial appearance features of objects interacted with as well as, on the other 
hand, the global context of the surrounding scene contain crucial cues. The paper presents a newly
designed CNN to extract such information by identifying informative image patches. 

\begin{figure*}[ht]
 \centering
 \includegraphics[width=500pt]{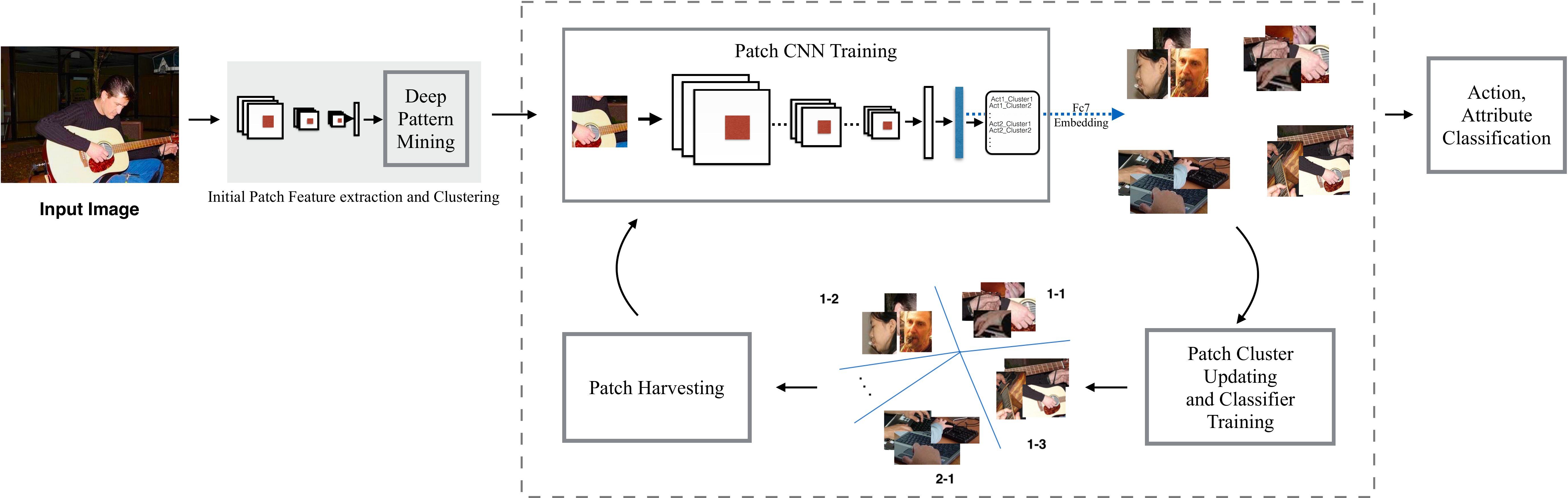}
 \caption{Full Pipeline of the proposed method for training mid-level deep visual elements in action and attribute. All the modules are explained in Sec.~\ref{sec:proposed}. The first box, which is the baseline of our work, initially cluster patches. The second box propose the introduced iterative process, and contains 3 main blocks. The final block takes trained classifiers and patch features of the second box after convergence, and classify images based on their action or attribute.}
 
 \label{fig:2}
 \end{figure*}

The idea of focusing on patches or parts definitely is not new in computer vision, also not when 
it comes to human actions or attributes \cite{EPM,Stanford40}. \cite{EPM} show that a good solution to human action classification can be achieved without trying to obtain a perfect pose estimation and without using
body part detectors. Indeed, an alternative is to capture discriminative image patches. Mining such patches for the cases of actions and attributes is the very topic of this paper. After deriving some initial discriminative patch clusters for each category of action or attribute,
our deep pattern CNN puts them into an iterative process that further optimizes the discriminative 
power of these clusters. Fig.~\ref{fig:2} sketches our CNN and will be explained further in the 
upcoming sections. At the end of the training, the CNN has become an expert in detecting those
image patches that distinguish human actions and attributes. The CNN comes with the features 
extracted from those patches. 

Our experiments show that we obtain better performance for action and attribute recognition than
top scoring, patch-based alternatives for object and scene classification~\cite{disc2, MDPM, maxmargin}. The latter do not seem to generalize well to the action and attribute case because these tasks need more fine-grained mid-level visual elements to make discrimination between similar classes. 

The rest of the paper is organized as follows. Related work is discussed in section \ref{sec:related}. 
Section~\ref{sec:proposed} describes our framework and new CNN for the mining and detection of discriminative patches for human action and attribute classification. Section~\ref{sec:exps}
evaluates our method and compares the results with the state-of-the-art. Section~\ref{sec:conclusion}
concludes the paper.

\section{Related Work}
\label{sec:related}
This section first discusses action and attribute recognition in the pre-CNN era. It then continues
with a short description of the impact that CNNs have had in the action and attribute recognition 
domain. Finally, we focus on the mid-level features that this paper shows to further improve 
performance. 

\paragraph{Action and Attribute Recognition.} Action and attribute recognition has been approached
using generic image classification methods~\cite{rel1,rel2,Spatialpyramid}, but with visual features extracted from human 
bounding boxes. Context cues are based on the objects and scene visible in the image, e.g. the mutual
context model \cite{contextrel4}. The necessary annotation of objects and human parts is substantial. 
Discriminative part based methods like DPM~\cite{DPM} have been state-of-the-art for quite a while. Inspired by their performance, human poselet methods \cite{poselet1,Berkeley_Attributes} try to capture ensembles of body and object parts in actions and attributes. Maji et al. \cite{Poselets} trained dedicated poselets for each
action category. In the domain of attributes the work by Parikh et al.~\cite{parikh2011relative} has
become popular. It ranks attributes by learning a function to do so. Berg et al.~\cite{bergAtt} proposed automatic attribute pattern discovery by mining unlabeled text and image data sampled 
from the web. Thus, also before the advent of CNNs some successes had been scored.

\paragraph{CNN powered Approaches.} Convolutional neural networks (CNN) have since defined the 
state-of-the-art for many tasks, incl. image classification and object detection \cite{alexnet,lecunCNN,RCNN,fastRCNN}. Many researchers proposed new CNN architectures or innovative 
methods on top of a CNN. Girshick et al.~\cite{RCNN} proposed a novel state-of-the-art object 
detection scheme (RCNN) by extracting CNN features from object region proposals. Gkioxari et 
al.~\cite{gkioxari2014r,RCNNstar} used a scheme similar to RCNN for action classification and 
detection, and for pose estimation. Zhang et al. \cite{PANDA} used HOG-poselets to train a CNN-based
part-based model for attribute classification. They achieved a nice gain over previous work. 
There also is recent work that trains models based on parts and poses \cite{birdCNN, 
RCNN*}. Zhang et al. \cite{birdCNN} obtained
a good performance with a part-based RCNN for bird species classification. The part-based RCNN can discriminate birds by learned models of their parts, by fine-tuning a CNN trained on ImageNet.
\cite{deepposelet} trained a deep CNN with prepared HOG poselets as training data and 
detected humans based on the resulting deep poselets. Recently Gkioxari et al.~\cite{RCNN*}
proposed to train human body part detectors, e.g. for the head and torso, based on CNN pool5 feature
sliding window search and combined them with the whole body box to train a CNN jointly. They showed 
that for the task of action and attribute classification, performance can be improved by adding 
such deep body part detectors to the holistic CNN. This work therefore suggests that adding 
dedicated patch analysis is beneficial. 

\paragraph{Discriminative Mid-level feature learning} Mid-level visual learning aims at capturing
information at a level of complexity higher than that of typical “visual words”. Mining visual 
elements in a large number of images is difficult since one needs to find similar discriminate 
patterns over a very large number of patches. The fine-grained nature of our action and attribute
tasks further complicates this search. \cite{disc1,disc2,disc3,disc4} describe methods for extracting 
clusters of mid-level discriminative patches. Doersch et al.~\cite{disc2} 
proposed such a scheme for scene classification, through an extension of the mean-shift algorithm 
to evaluate discriminative patch densities. Naderi et al.~\cite{naderiICLR} introduce a method to learn part-based models for scene classification which a joint training alternates between training part weights and updating parts filter.
One of the state-of-the-art contributions in mid-level 
element mining is \cite{MDPM}, which applies pattern mining to deep CNN patches. We have been 
inspired by the demonstration that the mining improves results. 

To the best of our knowledge, 
the use of CNN mid-level elements for action and attribute classification, as is the case in this
paper, is novel. Moreover, given the fine-grained nature of these challenges, we propose a new 
method to get more discriminative mid-level elements. The result is a performance better than 
that of competing methods.

\section{Method}
\label{sec:proposed}

In this section, we go through all our new framework for finding discriminative patch clusters and also our convolutional neural network for precise describing of patches.
In the first part of this section we talk about the motivation and give an overview of solution. Second part describes our proposed pipeline of mid-level patch mining and its contained blocks. Third part of the section introduces our proposed deep convolutional network for patch learning and the idea behind it. And in the final part we summarize that how we use mid-level visual elements in actions and attributes class-specific classifiers.

\subsection{Approach overview}

\begin{figure*}[ht]
 \centering
 \includegraphics[width=500pt]{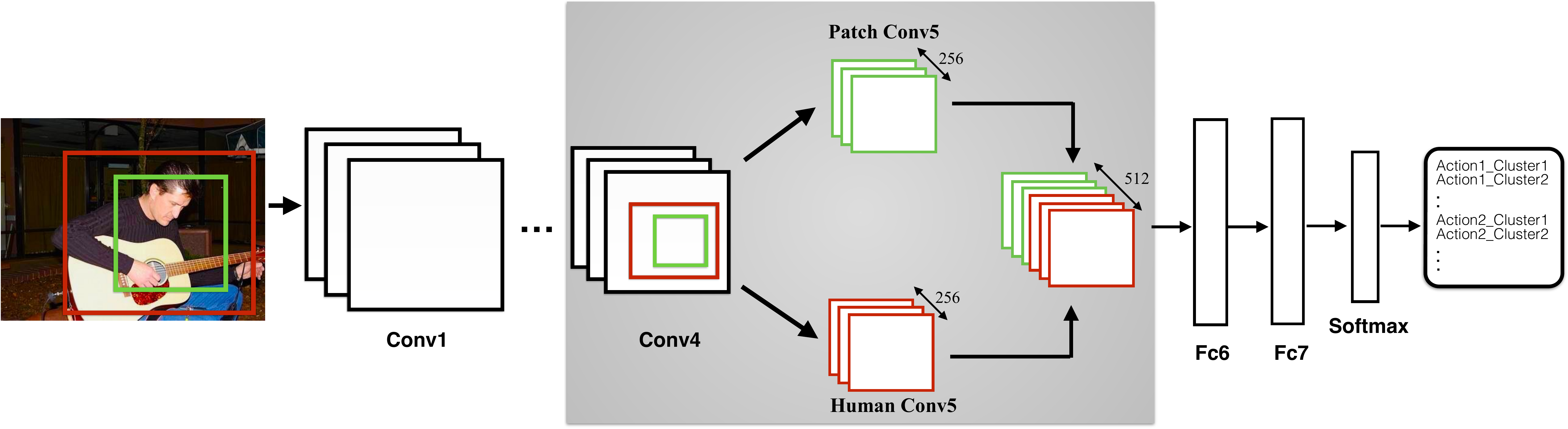}
 \caption{Overview of proposed Mid-level Deep Pattern Network. To train this CNN for mid-level discriminative patches, we concatenate the {\em conv5} layers of patch and the person regions to abstract the visual distinctive information of the patch with holistic clue of the person who is performing an action or has a specific attribute.}
 
 \label{fig:3}
 \end{figure*}

\hamed{
We address an approach using  mid-level deep visual patterns for actions and attributes classification which are fine-grained classification tasks. Applying discriminative patches or mid-level pattern mining state-of-the-arts like \cite{disc2,MDPM} to these tasks can not perform very promising as much as in the more generic classification tasks like scenes or object recognition (as we show in the experiments Sec \ref{sec:exps}).
The pattern mining algorithm \cite{MDPM} maps all data points to an embedding space to performs the association rule based clustering. For the embedding space, it fine-tunes AlexNet \cite{alexnet} for action or attribute recognition and uses its {\em fc7} layer to extract deep feature embedding. Our main insight in this paper is that a better embedding can improve the quality of clustering algorithm. We design an iterative algorithm where in each iteration, we improve the embedding by training a new CNN to classify cluster labels obtained in the previous iteration. In addition, we believe that aggregating the information and context from whole human body with specific action or attribute label with patches can improve the clusters of mid-level elements. Hence, we modify the architecture of AlexNet to concatenate features from both patch and the whole human bounding box in an intermediate layer (Fig.\ref{fig:3}). We show that learning the embedding using this new architecture outperforms the original AlexNet fine-tuned using patch images alone. Moreover, in each iteration, we purify the clusters by removing the patches that are scored poorly in the clustering. Subsequently, to classify actions and attributes by discriminative patches, we use a similar representation in \cite{MDPM,midObj} which more details about it come in Sec \ref{sec:finalClassify}. In the next part, we reveal more about the components of our pipeline. Finally, we show that the newly learned clusters produce better representations that outperform state-of-the-art when used in human action and attribute recognition. Our contributions are two-fold: (1) designing an iterative algorithm contains an expert patch CNN to improve the embedding, (2) proposing new patch CNN architecture training to use context in clustering the patches.
}


\subsection{Pipeline Detailes}
\label{sec:Blocks}

\hamed{
As shown both in Fig.\ref{fig:2} and Algorithm.\ref{alg1}, our iterative algorithm consists of four blocks which are described in more details in this section.
}

\ignore{
As it can be inferred from the Figure \ref{fig:2} our proposed pipeline consists of four main blocks. In this section we describe each of the blocks of pipeline with more details.
}
\paragraph{Initial feature extraction and clustering}
\hamed{
The first block clusters image patches discriminatively using Mid-Level Deep Pattern Mining (MDPM) algorithm \cite{MDPM}. Given, a set of training images annotated with humans' actions and their bounding boxes, it extracts a set of patches from the person bounding box and learns clusters that can discriminate between actions.  The MPDM method, building on the well-known association rule mining which is a popular algorithm in data mining, proposes a pattern mining algorithm, to solve mid-level visual element discovery. This approach in MDPM makes it interesting method because the specific properties of activation extracted from the fully-connected layer of a CNN allow them to be seamlessly integrated with association rule mining, which enables the discovery of category-specific patterns from a large number of image patches . They find that the association rule mining can easily fulfill two requirements of mid-level visual elements, representativeness and discriminativeness. After defining association rule patterns, MDPM creates many mid-level elements cluster based on shared patterns in each category and then applying their re-clustering and merging
algorithm to have discriminative patch cluster. We use the MDPM block to have initial mid-level elements clusters to move further on our method.
}

\paragraph{Training patch clusters CNN}
\hamed{
Our main insight is that the representation of image patches plays an important role in clustering. Assuming that the initial clustering is reasonable, in this block, we train a new CNN to improve the representation. The new CNN is trained so that given patch images, it predicts their cluster label. This is in contrast to the initial CNN that was learned to classify bounding box images to different action categories. We believe learning this fine-grained classification using discriminative patch cluster CNN results in a better representation for clustering.
}  

\newcommand{\nosemic}{\SetEndCharOfAlgoLine{\relax}}
\newcommand{\dosemic}{\SetEndCharOfAlgoLine{\string;}}
\newcommand{\pushline}{\Indp}
\newcommand{\popline}{\Indm\dosemic}
\begin{algorithm}
\caption{Iterative mid-level deep pattern learning.}\label{alg}

\nosemic
 \textbf{Input:} Image set ($\textbf{I}$ , $\textbf{L}$) \;
 Extract dense patche: $\textbf{P}^i_j$ (jth patch of ith image) \;
  Extract initial features $\textbf{F}^i_j$ and initial cluster labels $\textbf{C}^i_j$ \;
\While{Convergence}{ 
\textbf{$CNN_{Patch}$} = Train\_CNN($\textbf{P}$,$\textbf{C}$) \;

$\textbf{F} \leftarrow$ Extract\_CNN\_Feature($CNN_{Patch}$,$\textbf{P}$) \;
$\textbf{C} \leftarrow$ Update\_Cluster($\textbf{F}$,$\textbf{C}$) \;
$\textbf{W}$ = Train\_Patch\_Classifier($\textbf{F}$,$\textbf{C}$) \;
$\textbf{S}$= Compute\_Score($\textbf{W}$,$\textbf{F}$)  \;

\For {all patches}{

\If {$S^i_j<th$}{
Eliminate $P^i_j$}
}
}
\textbf{Output:} Mid-Level Pattern Clusters ({$\textbf{C}$})
\label{alg1}

\end{algorithm}

\ignore{
In this block we train our patch clusters convolutional network, with new inputs and outputs to solve the problem of feature extraction in the initial block. Densely extracted patches from images are inputs of this network, and the outputs are the computed sub-category labels by the initial clustering of MDPM block or the sub-category labels, which computed in the previous step of the iteration. As we propose a new network in this block, more details of this block described in the section \ref{sec:patchCNN}.
}

\paragraph{Updating clusters}
\hamed{
Now that a representation is learned by a newly trained CNN, we can update the clusters again using MDPM to get a better set of clusters that match the new representation. Since populating mid-level clusters in MDPM is time consuming, we freeze the first level of clustering and update the clusters by repeating re-clustering and merging using the new representations. This results in better clusters. Finally, we train new set of LDA classifiers to detect the clusters. The modification to MDPM to do re-clustering is described in Section \ref{ExperimentalSetup}.
}

\ignore{
The consequence of network training in the previous block is modification of feature space. Therefore we have updated features vectors for all of the instances. Using previous sub-class labels in clustering of these new features reduces the positive effects of training the convolutional network. Wherefore updating clusters is needed. On the other hand if we use a common clustering algorithm it may harm our previous knowledge about sub-clusters, which were based on finding simultaneously dense and discriminative clusters. To solve the issue, we just use cluster updating and merging of MDPM to obtain updated clusters of  extracted features or embedding from patches. The changes on MDPM block to do re-clustering described in Experiments section \ref{ExperimentalSetup}. By updating clusters we need a new classifier, therefore we use LDA, same as in MDPM block, to train classifiers on the updated features and clusters. 
}

\paragraph{Harvesting patches}
\hamed{
In order to improve the purity of clusters, we clean the clusters by removing patches that do not fit well in any cluster. We do this by thresholding the confidence value that LDA classifiers produce for each cluster assignment. Finally, we pass the new patches with associate cluster labels to learn a new CNN based representation. In the experiments, do cross validation, and stop the iterations when the performance on the validation set stops improving.
}

\subsection{Mid-level Deep Patterns Network}
\label{sec:patchCNN}
\hamed{
In updating the representation, we train a CNN to predict the cluster labels for given image patches. This is a challenging task for the network since clusters are defined to be action or attribute specific, so they can discriminate between actions. However, the patch image may not have enough information to discriminate between actions. Hence, to increase the discrimination power of the representation, we modify the network architecture to add the human bounding box image as extra contextual information in the network input (Fig. \ref{fig:3}). Following AlexNet architecture, we pass both patch image and the whole bounding box image to the network and concatenate the activations in {\em conv5} layer to form a larger convolutional layer. To train our mid-level deep patterns CNN, we try fast RCNN \cite{fastRCNN}. In training process of fast RCNN for patch learning, we push two region: patch and the croped image of person.  An adaptive max pooling layer takes the output of the last convolutional layer and a list of regions as input. We concatenate the ROI-pooled {\em conv5} features from two regions and then pass this new {\em conv5}(concatenated) through the fully connected layers to make the final prediction. Using fast RCNN helps us to have an efficient, fast and computationally low cost CNN layers calculations, since convolutions are applied at an image-level and are afterward reused by the ROI-specific operations. Our network is using a pre-trained CNN model on ImageNet-1K with Alex-Net\cite{alexnet} architecture, to perform fine-tuning and learn patch network.
}

\subsection{Action and attribute classifiers}
\label{sec:finalClassify}
\hamed{
After learning the mid-level pattern clusters, we use them to classify actions and attributes. Given an image, we extract all patches and find the best scoring one for each cluster. To construct the image representation for action or attribute on each image, we use the idea of mid-level elements for object detection \cite{midObj}, by taking the max score of all patches per mid-level pattern detectors per region encoded in a 2-level (1 * 1 and 2 * 2) spatial pyramid. This feature vector represents occurrence confidence of elements in the image. This results in a rich feature for action and attribute classification since the clusters are learned discriminatively for this task. Finally, we pass the whole bounding box through overall CNN trained on action or attribute labels and append it's {\em fc7} activations to obtained feature.
}

\section{Experiments}
\label{sec:exps}
We evaluate our algorithm on two tasks: action classification and attribute classification in still images. In both tasks, we are folowing the stantdard PASVAL VOC \cite{pascal} setting that the human bounding box is given in the inference time. The first section of our evaluations are on the PASCAL VOC
\cite{pascal}
and Stanford 40  
\cite{Stanford40}
action datasets, and the second part is on the Berkeley attributes of people dataset
\cite{Berkeley_Attributes}.

\subsection{Experimental Setup}
\label{ExperimentalSetup}
\paragraph{Common properties of the networks}
All of the networks have been trained using the caffe CNN training package
\cite{caffe}
with back-propagation. We use weights of the trained Network on ImageNet dataset
\cite{imagenet}
 as initial weights and fine-tune our networks on specific datasets and with different properties according to the task. We set the learning rate of CNN training to 0.0001, the batch size to 100.

\paragraph{Initial feature extraction network}
The fine-tuning of the network is done on the cropped images of each person as input and the Action or Attribute label of images as output of the network. Then we use {\em fc7} feature vector of body image or extracted patches as input of the MDPM (Mid-level Deep Pattern Mining) \cite{MDPM} block.

\paragraph{Mid-level deep patterns network}
Input images of this network are patches that extracted from cropped body image in 3 different scales (128*128, 160*160, 192*192 patches from a resized image with stride of 16). The output layer of this network is cluster labels that computed by MDPM block. 

\begin{table*}[t]
  \centering
  \resizebox{\textwidth}{!}{
  \begin{tabular}{  c | c c c c c c c c c c | c } 

 AP(\%) & Jumping & Phoning & Playing Instrument & Reading & Riding Bike & Riding Horse & Running & Taking Photo & Using Computer & Walking & mAP\\ [0.5ex]
 \hline

CNN & 76.2 &46.7 &75.4 &42.1 &91.4 &93.2 &79.1 &52.3 &65.9 &61.8 & 68.4\\ [0.5ex]

CNN+MDPM & 76.8 &47.7 &75.6 &44.1 &90.4 &93.8 &80.1 &53.6 &65.4 &62.7 & 69\\ [0.5ex]

Ours\_AlexNet\_iter1 & 76.9 &48.2 &74.9 &46.8 &91.6 &93.9 &82.1 &54.3 &66.4 &63.5 & 69.9\\[0.5ex]

Ours\_AlexNet\_iter2 & 78.5 &49.3 &77.9 &50.2 &92.1 &94.2 &82.4 &56.4 &70.1 &64.3 & 71.5\\[0.5ex]

Ours\_AlexNet\_iter3 & 78.1 &49.8 &77.8 &51.2 &92.1 &94.6 &82.7 &56.5 &70.3 &64.2 & 71.7\\[0.5ex]

\hline

Ours\_PatternNet\_iter1 & 80.1 &53.7 &78.3 &55.2 &93.2 &94.8 &84.7 &57 &72.2 &66.2 & 73.5\\ [0.5ex]

Ours\_PatternNet\_iter2 & 81.2 &55.4 &80.1 &60.1 &94.3 &95.1 &86.7 &59.1 &73.3 &67.8 & 75.3\\ [0.5ex]

\textbf{Ours\_PatternNet\_iter3} & 81.4 &55.3 &80.3 &60.3 &95 &94.8 &86.2 &59.4 &73.6 &68 & \textbf{75.4}\\ [2ex]

\end{tabular}
}
  \caption{Average Precision on the PASCAL VOC dataset validation set. The two first rows are baselines of our method, which are results of training CNN on pascal and using MDPM to classify them. The ours\_Alex methods rows are the results of iterating 1,2 and 3 times in the iterative process of pipeline using the Alex-Net architecture as patch CNN training block. The Ours\_PatternNet are same as previous ones by using our proposed mid-level deep patterns network.}
  \label{tab:1}
\end{table*}

\begin{table*}[t]

  \centering
  \resizebox{\textwidth}{!}{
  \begin{tabular}{  c | c c c c c c c c c c | c } 

 AP(\%) & Jumping & Phoning & Playing Instrument & Reading & Riding Bike & Riding Horse & Running & Taking Photo & Using Computer & Walking & mAP\\ [0.5ex]
 \hline

   CNN & 77.1 &45.8 &79.4 &42.2 &95.1 &94.1 &87.2 &54.2 &67.5 &68.5 & 71.1\\ [0.5ex]
  
CNN+MDPM & 77.5 &47.2 &78.3 &44.2 &94.2 &95.3 &89.2 &56.4 &68.1 &68.3 & 71.9\\ [0.5ex] 

 Action Poselets & 59.3 & 32.4 & 45.4 & 27.5 & 84.5 & 88.3 & 77.2 & 31.2 & 47.4 & 58.2 & 55.1\\ [0.5ex] 
  
Oquab et al & 74.8 & 46.0 & 75.6 & 45.3 & 93.5 & 95.0 & 86.5 & 49.3 & 66.7 & 69.5 & 70.2 \\ [0.5ex]

Hoai& 82.3 & 52.9 & 84.3 & 53.6 & 95.6 & 96.1 & 89.7 & 60.4 & 76.0 & 72.9 & 76.3\\ [0.5ex]

Gkioxari et al & 77.9 & 54.5 & 79.8 & 48.9 & 95.3 & 95.0 & 86.9 & 61.0 & 68.9 & 67.3 & 73.6\\[0.5ex]
  
\hline

Ours\_AlexNet & 79.6 &51.7 &79.7 &50.8 &94.6 &95.8 &88.9 &58.4 &71.1 &68.7 & 73.9\\ [0.5ex]

\textbf{Ours\_PatternNet} & 81.4 & 53.8 & 86 & 54.9 & 96.8 & 97.5 & 91.4 & 62.1 & 78.0  & 74.5 & \textbf{77.6}\\ [2ex]

\end{tabular}
}
  \caption{Average Precision on the PASCAL VOC dataset test set and comparison with previous methods. The first two rows are our baselines which reported on the test set, the next rows of the above part are previous methods based on 8 layer convolutional network, same as ours. The ours\_Alex and ours\_PatternNet are the results of testing our proposed pipeline with Alex-Net and our Pattern-Net architectures, on the test set of PASCAL VOC, until the convergence of iteration (3 iterations).}
  \label{tab:2}
\end{table*}

\paragraph{Mid-level deep pattern mining block.}
We use MPDM block with the mentioned properties in \cite{MDPM} for the initial feature extraction and clustering block. 
While updating clusters in our iterative patch clusters training, we use a part of MPDM algorithm which tries to merge and reconfigure clusters. The new obtained CNN representations for patches help updating clusters to be performed more precise. We apply MDPM patch mining with 50 cluster per each category.

 
\subsection{Action Classification.}
For the action Classification task, we evaluate our mid-level pattern mining pipeline and proposed patch CNN network performances on PASCAL VOC and Stanford 40 action datasets.

\paragraph{Dataset.}
The PASCAL VOC action dataset \cite{pascal} includes 10 different action classes including Jumping, Phoning, Playing Instrument, Reading, Riding Bike, Riding Horse, Running, Taking Photo, Using Computer, Walking, and an Other class consists of images of persons, which has no action label. The dataset has 3 splits of training, validation and test set.
~\\~\\
The Stanford 40 action dataset \cite{Stanford40} contains total of 9532 images and 40 classes of actions, split into train set containing 4000, and test set containing 5532 instances. 

\paragraph{Implementation detailes.}
The training and fine-tuning of the initial CNN and pattern CNN,  have been done only on PASCAL VOC dataset. It means to evaluate on Stanford40, we just use convolutional networks of action and patch clusters, which are trained on PASCAL and afterward run the MDPM cluster mining and configure clusters for Stanford40.

In the test time we will evaluate the results on both PASCAL VOC and Stanford 40 datasets. The reason of training patch CNN networks on a dataset with less classes than the test dataset is to evaluate discrimination power of our proposed method's extracted patches. In the results section we show that our method achieves state-of-the-art on the both of PASCAL VOC and Stanford 40 dataset, which consequently with results on Stanford40, the discrimination power of extracted patches has been proved.
\begin{table}
\begin{center}
\begin{tabular}{|l|c|}
\hline
Method & AP(\%) \\
\hline\hline
Object bank & 32.5 \\
LLC & 35.2 \\
SPM & 34.9 \\
EPM & 40.7 \\
CNN\_AlexNet & 46\\
CNN+MDPM & 46.8\\
\hline
Ours\_AlexNet& 49\\
\textbf{Ours\_PatternNet}& \textbf{52.6}\\
\hline
\end{tabular}
\end{center}
\caption{Average Precision on the Stanford40 action dataset. The used initial CNN and patch CNNs in this section are trained on the PASCAL VOC dataset, and we use these networks to extract patches form Images of Stanford40 dataset.}
  \label{tab:4}
\end{table}

\begin{table*}[t]

\centering
\resizebox{\textwidth}{!}{
\begin{tabular}{  c | c c c c c c c c c | c } 

AP(\%) & is male & has long hair & has glasses & has hat & has t-shirt & has long sleeves & has shorts & has jeans & has long pants & mAP\\ [0.5ex]
\hline

CNN$_{att}$ & 88.6 &82.2 &50.1 &83.2 &60.1 &86.2 &88.3 &88.6 &98.2 & 80.6 \\ [0.5ex]
CNN$_{att}$+MDPM & 88.8 &84.2 &54.1 &83.4 &64.3 &86.4 &88.5 &88.8 &98.3 & 81.9 \\ [0.5ex]
PANDA & 91.7 &82.7 &70 &74.2 &49.8 &86 &79.1 &81 &96.4 & 79 \\ [0.5ex]
Gkioxari et al & 91.7 &86.3 &72.5 &89.9 &69 &90.1 &88.5 &88.3 &98.1 & 86\\ [0.5ex]
\hline
Ours\_AlexNet & 90.8 &84.2 &61.4 &88.9 &67.1 &88.1 &89.2 &89.3 &98.3 & 84.1\\ [0.5ex]
\textbf{Ours-PatternNet} & 91.8 &88.4 &71.1 &88.9 &70.7 &91.8 &88.7 &89.3 &98.9 & \textbf{86.6}\\ [2ex]

\end{tabular}
}
\caption{Average Precision on the Berkeley Attributes dataset and comparison with previous methods. The CNN$_{att}$ and CNN$_{att}$+MDPM are the baselines of the work, which their convolutional networks trained on train set of Berkeley attributes dataset. The results of PANDA method with 5 layer network and 8 layer network results of Gkioxari et al is reported in last rows of above part. The bottom of the table shows the results of our proposed pipeline using both Alex-Net and Our Patch-Net until the convergence of the iteration process (3 iterations).}
\label{tab:3}
\end{table*}

 \begin{figure*}[ht]

\resizebox{18cm}{!}{
\includegraphics[width=500pt]{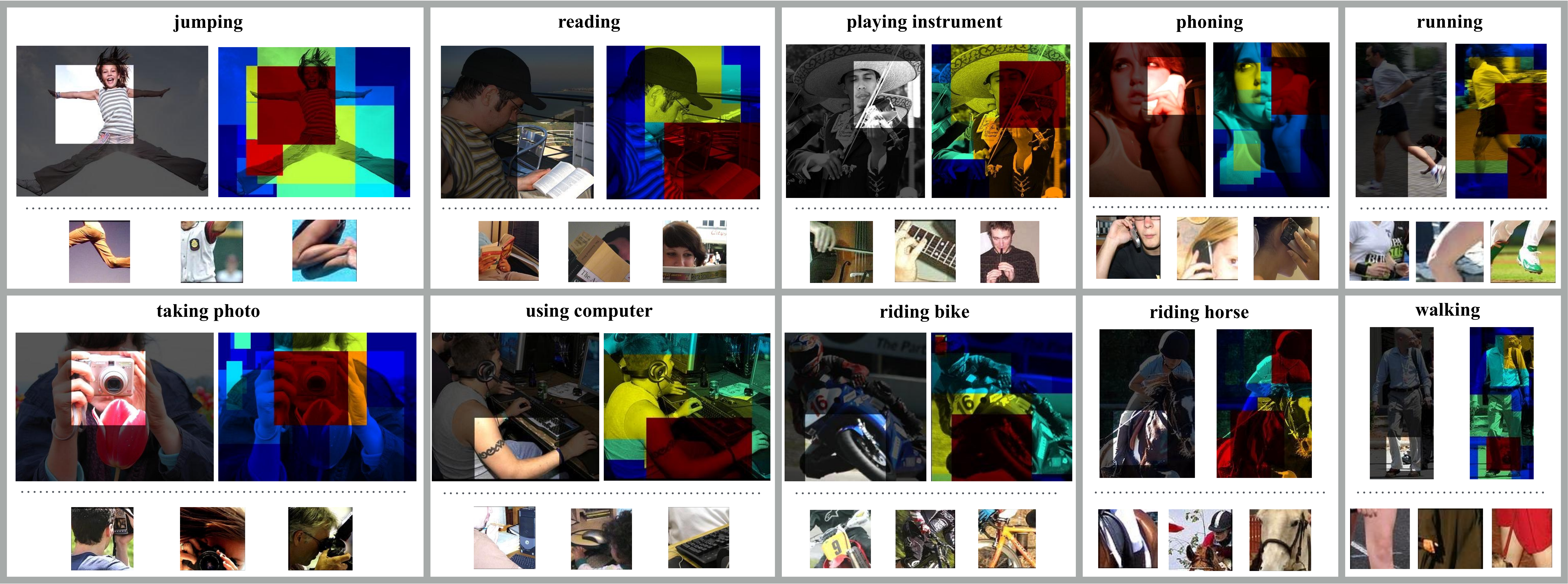}

 \label{fig:4}
 } 
 \caption{Explored deep mid-level visual patterns of different categories of actions and samples detected from top scored pattern and aggregated scores over all image from PASCAL VOC 2012 action dataset. In each block of figure, small patches are representatives from most discriminative patches.}
 
 \end{figure*}

\paragraph{Results.}
We report the result of our baseline, and proposed method on the PASCAL VOC validation set in  Table \ref{tab:1}. The baseline 'CNN' in the first row of table is AlexNet trained on PASCAL VOC dataset using SVM on the {\em fc7} layer features. The second row which is the output of our initial feature extraction and clustering block, named 'CNN+MDPM' reports the result of SVM training on the concatenated feature vector of convolutional network {\em fc7} and the 2500 dimensional feature vector output of MDPM block (50 cluster * 10 category * 5 spatial pyramid region). The next three rows of the table with names of 'Ours\_AlexNet\_iter1-3' show the result of performing the pipeline using the convolutional neural network architecture of AlexNet and with 1 to 3 iterations. Finally the last three rows are same as previous ones with 1 to 3 iterations applying our proposed pattern CNN architecture. We can conclude from the table that our proposed iterative pipeline and newly proposed CNN architecture can improve the result independently, so the combined method outperform either one alone.

The results of our final proposed method in comparison with results of the following methods, Poselets
\cite{Poselets}
, Oquab et al
\cite{oquab2014}
,Hoai 
\cite{hoai2014}
, and Gkioxari et al 
\cite{RCNN*} 
on the test set of PASCAL VOC has been shown in Table \ref{tab:2}. As we can see in the table the mean accuracy of our proposed method with the proposed PatternNet outperforms all the previous 8 layer CNN network based methods. The important point in this improvement in result is that most of the mentioned methods were using part detectors based on the part and pose annotations of the datasets which limits the number of annotated training data because of the hardness of pose annotating. In contrast the proposed method does not use any annotation more than action labels and bounding box of person. 

As we mentioned in the implementation details we evaluate the Stanford40 actions dataset using our final pipeline mid-level patterns CNN - PatternNet\_iter3 - which is trained on PASCAL VOC, and report the results in Table \ref{tab:4}. The result shows that our method improved the results of the previous methods in action classification on this human action dataset.

\subsection{Attribute Classification}
In this section we report implementation details and results of our method on the Berkeley Attributes of people dataset. We need to train all the networks on the new dataset.

\paragraph{Dataset.}
The Berkeley attributes of people dataset contains 4013 training and 4022 test examples of people, and 9 Attributes classes, is male, has long hair, has glasses, has hat, has t-shirt, has long sleeves, has shorts, has jeans, has long pants. Each of the classes labeled with 1,-1 or 0, as present, absent and un-specified labels of the attribute. 
\paragraph{Implementation details.}
\label{attr_implement}
In contrast to action classification task in attributes classification, more than one label can be true for each instance, it means that classes in attribute classification do not oppose each other. Therefore instead of using the softmax function as the loss function in the last layer of the initial convolutional network, which forces the network to have only a true class for each instance, we use cross entropy function for the task of attributes classification. 

The other block with the same assumption in opposition of classes is MDPM block which try to find some cluster for each class such that instances of other classes labels as negative to maximize the discrimination of clusters. In the other hand, attribute classes do not oppose each other, so a modification is needed in the MDPM block. We extract discriminative clusters of each class separately, using the positive and negative labels of that class.

\paragraph{Results.}
We evaluate our method on the Berkeley attributes of people dataset and compare our results on the test set with Gkioxari et al
\cite{RCNN*}
and PANDA
\cite{PANDA} methods in Table \ref{tab:3}. As we show in the table, our baselines, 'CNN$_{att}$' and 'CNN$_{att}$ +MDPM' did not improve the results of previous methods. Even our proposed pipeline with the AlexNet architecture couldn't outperform \cite{RCNN*} which uses trained deep body parts detectors. However, our proposed pipeline with the proposed PatternNet architecture improves the result of attribute classification in comparison with all previous methods. Table \ref{tab:3} shows that although our method have significant improvements in action classification, the method does not have the same margin with the state-of-the-arts in classifying attributes. We believe this is due to the importance of part annotations in training attribute classifier, which is not available in our setting. 
\section{Conclusion}
\label{sec:conclusion}
In this work, we have addressed human action and attribute classification using mid-level discriminative visual elements. We proposed a novel framework to learn such elements using a Deep Convolutional Neural Network which also has a new architecture. The algorithm explores a huge number of candidate patches, covering human body parts as well as scene context.  We validated our method on the PASCAL VOC 2012 action, the Stanford40 actions, and the Berkeley Attributes of People datasets.
The results are good, both qualitatively and quantitatively,  reaching the state-of-the-art, but without using any human pose or part annotations.

\subsection*{Acknowledgements}
This work was supported by DBOF PhD scholarship, KU Leuven CAMETRON project.

{
\small	
\bibliographystyle{ieee}
\bibliography{egbib}
}

\end{document}